\documentclass{article}

\usepackage{microtype}
\usepackage{graphicx}
\usepackage{subfigure}
\usepackage{booktabs} 

\usepackage{hyperref}

\usepackage[accepted]{icml2019}

\usepackage{amsmath,amssymb,amsfonts}
\usepackage{textcomp}
\usepackage{diagbox}
\usepackage{multirow}
\usepackage{caption}
\usepackage{url}
\usepackage{etoolbox}
\usepackage{makecell}
\usepackage{cleveref}
\usepackage{soul}
\newcommand{\argmin}{\operatornamewithlimits{argmin}}
\newcommand{\norm}[1]{\left\lVert#1\right\rVert}

\newif\ifrepurpose
\repurposefalse

\newif\ifztable
\ztablefalse

\newif\iffederated
\federatedfalse

\newif\iftrimmed
\trimmedfalse

\newif\ifrparams
\rparamsfalse

\begin{document}
\twocolumn[
\icmltitle{A Little Is Enough: Circumventing Defenses For Distributed Learning }

\icmlauthor{Moran Baruch}{biu}
\icmlauthor{Gilad Baruch}{biu}
\icmlauthor{Yoav Goldberg}{biu}
\icmlaffiliation{biu}{Dept. of Computer Science, Bar Ilan University, Israel}
\icmlcorrespondingauthor{Moran Baruch}{moran.baruch@biu.ac.il}
\icmlcorrespondingauthor{Gilad Baruch}{gilad.baruch@biu.ac.il}

\vskip 0.3in
]
\printAffiliationsAndNotice{} 

\begin{abstract}
Distributed learning is central for large-scale training of deep-learning models. However, they are exposed to a security threat in which Byzantine participants can interrupt or control the learning process. Previous attack models and their corresponding defenses assume that the rogue participants are (a) omniscient (know the data of all other participants), and (b) introduce large change to the parameters. We show that small but well-crafted changes are sufficient, leading to a novel \textit{non-omniscient} attack on distributed learning that go undetected by all existing defenses. We demonstrate our attack method works not only for preventing convergence but also for repurposing of the model behavior (``backdooring''). We show that 20\% of corrupt workers are sufficient to degrade a CIFAR10 model's accuracy by 50\%, as well as to introduce backdoors into MNIST and CIFAR10 models without hurting their accuracy.

\end{abstract}

\icmlkeywords{
distributed learning, adversarial machine learning, secure cloud computing.
}

\section{Introduction}

\textit{Distributed Learning} has become a wide-spread framework for large-scale model training \cite{dist1,PS1,PS2,pytorch,dist2, dist3, dist4}, in which a server is leveraging the compute power of many devices by aggregating local models trained on each of the devices. 

A popular class of distributed learning algorithms is \textit{Synchronous Stochastic Gradient Descent} (sync-SGD), using a single server (called \textit{Parameter Server} - PS) and $n$ workers, also called \textit{nodes} \cite{PS1, PS2}. In each round, each worker trains a local model on his or her device with a different chunk of the dataset, and shares the final parameters with the PS. The PS then aggregates the parameters of the different workers, and starts another round by sharing with the workers the resulting combined parameters to start another round. The structure of the network (number of layers, types, sizes etc.) is agreed between all workers beforehand.

While effective in sterile environment, a major risk emerge with regards to the correctness of the learned model upon facing even a single \textit{Byzantine} worker \cite{krum}. Such participants are not rigorously following the protocol either innocently, for example due to faulty communication, numerical error or crashed devices, or adversarially, in which the Byzantine output is well crafted to maximize its effect on the network. 

We consider malicious Byzantine workers, where an attacker controls either the devices themselves, or even only the communication between the participants and the PS, for example by \textit{Man In The Middle} attack. Both attacks and defenses have been explored in the literature \cite{krum, MeanMed, TrimmedMean, Bulyan, Auror}.

In the very heart of distributed learning lies the assumption that the parameters of the trained network across the workers are independent and identically distributed (i.i.d.) \cite{iid2, krum, TrimmedMean}. This assumption allows the averaging of different models to yield a good estimator for the desired parameters, and is also the basis for the different defense mechanisms, which try to recover the original mean after clearing away the byzantine values.
Existing defenses claim to be resilient even  when the attacker is omniscient \cite{krum, Bulyan, MeanMed}, and can observe the data of all the workers.  Lastly, all existing attacks and defenses \cite{krum, Bulyan, MeanMed, TrimmedMean} work under the assumption that achieving a  malicious objectives requires large changes to one or more parameters. This assumption is advocated by the fact that SGD \textit{better} converges with a little random noise \cite{noise, noise2, noise3}.

We show that this assumption is incorrect: directed \emph{small} changes to many parameters of few workers are capable of defeating all existing defenses and interfering with or gaining control over the training process.
Moreover, while most previous attacks focused on preventing the convergence of the training process, 
we demonstrate a wider range of attacks and support also introducing
\textit{backdoors} to the resulting model, which are samples that will produce the attacker's desired output, regardless of their true label. Lastly, by exploiting the i.i.d assumption we introduce a \emph{non-omniscient} attack in which the attacker only has access to the data of the corrupted workers.

\paragraph{Our Contributions}
We present a new approach for attacking distributed learning with the following properties:
\begin{enumerate}
\item We overcome all existing defense mechanisms.
\item We compute a perturbation range in which the attacker can change the parameters without being detected \textbf{even in i.i.d. settings}.
\item Changes within this range are sufficient for both interfering with the learning process \textbf{and for backdooring the system}.
\item We propose the first \textbf{non-omniscient} attack applicable for distributed learning, making the attack stronger and more practical. 
\end{enumerate}

\section{Background}
\subsection{Malicious Objectives}
\paragraph{Convergence Prevention} 
This is the attack which most of the existing attacks and defenses literature for distributed learning focuses on \cite{krum,Bulyan,MeanMed}. In this case, the attacker interferes with the process with the mere desire of obstructing the server from reaching good accuracy.
This type of attack is not very interesting because the attacker does not gain any future benefit from the intervention.
Furthermore, the server is aware of the attack and, in a real world scenario, is likely to take actions to mitigate it, for example by actively blocking subsets of the workers and observing the effect on the training process. 

\textbf{Backdooring}, also known as
\textbf{Data Poisoning} \cite{Poisoning,backdoor3, backdoor2}, is an attack in which the attacker manipulates the model at training time so that it will produce the attacker-chosen target at inference time. The backdoor can be either a single sample, e.g. falsely classifying a specific person as another, or it can be a class of samples, e.g. setting a specific pattern of pixels in an image will cause it to be classified maliciously.

\begin{figure}[thb]
\centering
\includegraphics[width=0.36\textwidth]{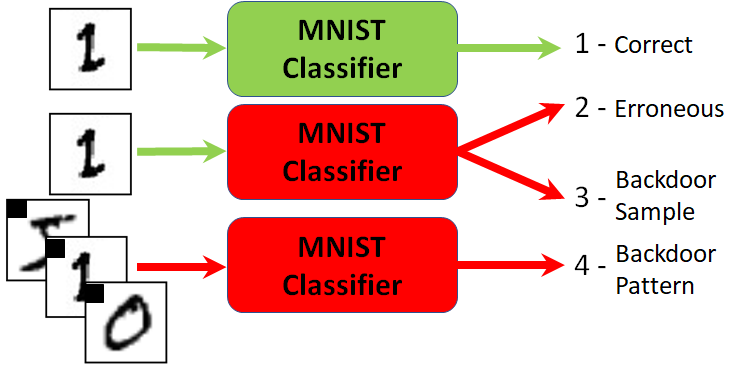}
\caption{\textbf{Possible Malicious objectives.} \textbf{1.} A normal scenario in which a benign image is classified  correctly. \textbf{2.} The malicious opponent damaged the network functionality which now mis-classify legitimate inputs. \textbf{3.} A backdoor appear in the model, classifying this specific image as the attacker desire. \textbf{4.} The model produces the label \textbf{4} whenever a specific pattern (e.g. a square in the top left) is applied.}\label{objectives}
\end{figure}

An illustration of those objectives is given in Figure~\ref{objectives}.

\subsection{Existing Attacks}
Distributed training is using the Synchronous SGD protocol, presented in Algorithm 
~\ref{alg:sync_sgd}.

\begin{algorithm} [h]
\begin{algorithmic}[1]
\STATE $P^1 \gets$ Randomly initiate the parameters in the server.
\FOR {round $t \in [T]$ }
    \STATE The server sends $P^t$ to all $n$ workers.
    \FOR {each worker $i \in [n]$}
        \STATE Set $P^t$ as initial parameters in the local model.
        \STATE Train locally using its own data chunk. \label{trainLocally}
        \qquad \qquad \quad $\lhd$ Malicious intervention
        \STATE Return final parameters $p^{t+1}_i$ to the server. \label{returnParams}
    \ENDFOR
    \STATE $P^{t+1} \gets AggregationRule(\{p_i^{t+1} : i \in [n]\}$)\; \label{lst:line:blah2}
    \STATE The server evaluates $P^{t+1}$ on the test set.
\ENDFOR
\STATE {\bfseries return} $P^t$ that maximized accuracy on the test set.
\caption {\sl Synchronous SGD}\label{alg:sync_sgd}
\end{algorithmic}
\end{algorithm}

The attacker interferes the process at the time that maximizes its effect, that is between lines~\ref{trainLocally} and \ref{returnParams} in Algorithm~\ref{alg:sync_sgd}. During this time, the attacker can use the corrupted workers' parameters expressed in $p_i^{t+1}$, and replace them with whatever values it desires to send to the server.
Attacks method differ in the way in which they set the parameter values, and defenses methods attempt to identify corrupted parameters and discard them.

Algorithm~\ref{alg:sync_sgd} aggregates the workers values using averaging ($AggregationRule()$ in line~\ref{lst:line:blah2}). Some defense methods change this aggregation rule, as explained below.  

\paragraph{Backdooring attacks}
Bagdasaryan et al. \yrcite{vitaly2018backdoor} demonstrated a backdooring attack on \emph{federated} learning by making the attacker optimize for a model with the backdoor while adding a term to the loss that keeps the new parameters close to the original ones. Their attack has the benefits of requiring only a few corrupted workers, as well as being non-omniscient. However, it does not work for distributed training: in federated learning each worker is using its own private data, coming from a different distribution, negating the i.i.d assumption \cite{fed1,fed2} and making the attack easier as it drops
the ground under the fundamental assumption of all existing defenses for distributed learning. \cite{sybils} proposed a defense against backdoors in federated learning, but like the attack above it heavily relies on the non-i.i.d property of the data, which does not hold for distributed training. 

A few defenses aimed at detecting backdoors were proposed \cite{backdoorDefense3, backdoorDefense4, backdoorDefense, backdoorDefense2}, but those defenses assume a single-server training in which the backdoor is injected in the training set for which the server has access to, so that by clustering or other techniques the backdoors can be found and removed. In contrast, in our settings, the server has no control over the samples which the workers adversely decide to train with, rendering those defenses inoperable.
Finally, \cite{Auror} demonstrate a method for circumventing backdooring attacks on distributed training. As discussed below, the method is a variant of the Trimmed Mean defense, which we successfully evade.

\subsection{Existing Defenses}
All existing defenses are working on each round separately, so for the sake of readability we will discard the notation of the round ($^t$).
For the rest of the paper we will use the following notations: $n$ is the total number of workers, $m$ is the number of corrupted workers, and $d$ is the number of dimensions (parameters) of the model. $p_i$ is the vector of parameters trained by worker $i$, $(p_i)_j$ is its $j$th dimension, and $\mathcal{P}$ is $\{p_i : i \in [n]\}$.

The state-of-the-art defense for distributed learning is \textit{Bulyan}. Bulyan utilizes a combination of two earlier methods - \textit{Krum} and \textit{Trimmed Mean}, to be explained first.

\paragraph{Trimmed Mean} This family of defenses, called \textit{Mean-Around-Median} \cite{MeanMed} or \textit{Trimmed Mean} \cite{TrimmedMean}, change the aggregation rule of Algorithm~\ref{alg:sync_sgd} to a trimmed average, handling each dimension separately:
\begin{equation}
TrimmedMean(\mathcal{P}) = \left\{v_j = \frac{1}{|{U_j}|}\sum_{i \in U_j}(p_i)_j : j \in [d] \right\}
\end{equation} 

Three variants exist, differing in the definition of $U_j$.
\begin{enumerate}
\item $U_j$ is the indices of the top-$(n-m)$ values in $\{(p_1)_j,$ $..., (p_n)_j\}$ nearest to the median  $\mu_j$ \cite{MeanMed}.
\item Same as the first variant only taking top-$(n-\textbf{2}m)$ values \cite{Bulyan}.
\item $U_j$ is the indices of elements in the same vector $\{(p_1)_j,...,(p_n)_j\}$ where the largest and smallest $m$ elements are removed, regardless of their distance from the median \cite{TrimmedMean}.
\end{enumerate}

A defense method of \cite{Auror} clusters each parameter into two clusters using 1-dimensional k-means, and if the distance between the clusters' centers exceeds a threshold, the values compounding the smaller cluster are discarded. This can be seen as a variant of the Trimmed Mean defense, because only the values of the larger cluster which must include the median will be averaged while the rest of the values will be discarded.

All variants are designed to defend against up to $\lceil \frac{n}{2} \rceil -1$ corrupted workers, as this defenses depend on the assumption that the median is taken from the range of benign values.

The circumvention analysis and experiments  are similar for all variants upon facing our attack, so we will consider only the second variant which is used in Bulyan below.

\paragraph{Krum} 
Suggested by Blanchard et al \yrcite{krum}, \textit{Krum} strives to find a single honest participant which is probably a good choice for the next round, discarding the data from the rest of the workers. The chosen worker is the one with parameters which are closest to another $n-m-2$ workers, mathematically expressed by:
\begin{equation}
Krum(\mathcal{P}) = \left( p_i \mid \argmin_{i \in [n]} \sum_{i \to j} \norm{p_i - p_j}^2 \right)
\end{equation}
Where $i \to j$ is the $n-m-2$ nearest neighbors to $p_i$ in $P$, measured by Euclidean Distance.

Like TrimmedMean, Krum is designed to defend against up to $\lceil \frac{n}{2} \rceil -1$ corrupted workers ($m$). 
The intuition behind this method is that in normal distribution, the vector with average parameters in each dimension will be the closest to all the parameters vectors drawn from the same distribution. By considering only the distance to the closest $n-m-2$ workers, sets of parameters which will differ significantly from the average vector are outliers and will be ignored. The malicious parameters, assumed to be far from the original parameters, will suffer from the high distance to at least one non-corrupted worker, which is expected to prevent it from being selected.

While Krum was proven to converge, in \cite{Bulyan} the authors already negate the proof that Krum is \textit{($\alpha$-f) Byzantine Resilient} (A term coined by Krum's authors), by showing that convergence alone should not be the target, because the parameters may converge to an \textit{ineffectual} model.
Secondly, as already noted in \cite{Bulyan}, due to the high dimensionality of the parameters, a malicious attacker can notably introduce a large change to a single parameter without a considerable impact on the L$^p$ norm (Euclidean distance), making the model ineffective.

\paragraph{Bulyan} 

El Mhamdi et al. \yrcite{Bulyan}, who suggested the above-mentioned attack on Krum, proposed a new defense that successfully oppose such an attack. They present a ``meta"-aggregation rule, where another aggregation rule $\mathcal{A}$ is used as part of it.
In the first part, Bulyan is using $\mathcal{A}$ iteratively to create a \textit{SelectionSet} of probably benign candidates, and then aggregates this set by the second variant of TrimmedMean.
Bulyan combines methods working with L$^p$ norm that proved to converge, with the advantages of methods working on each dimension separately, such as TrimmedMean, overcoming Krum's disadvantage described above because TrimmedMean will not let the single malicious dimension slip. 

Algorithm~\ref{bul_algo} describes the defense. It should be noted that on line~\ref{bul_meanmed}, $n-\bf{4}m$ values are being averaged, which is $n'-\bf{2}m$ for $n'=|SelectionSet|=n-2m$. 

\begin{algorithm}
\begin{algorithmic}[1]
\STATE {\bfseries Input:} $\mathcal{A}, \mathcal{P}, n, m$
\STATE $SelectionSet \gets \emptyset$
\WHILE{$|SelectionSet| < n-2m$}
    \STATE $p \gets \mathcal{A}(\mathcal{P} \setminus SelectionSet)$
    \STATE $SelectionSet \gets SelectionSet \cup \{p\}$
\ENDWHILE
\STATE \bf{return} $TrimmedMean_{(2)}(SelectionSet$) \label{bul_meanmed}

\caption{Bulyan Algorithm}\label{bul_algo}
\end{algorithmic}
\end{algorithm}

Unlike previous methods, Bulyan is designed to defend against only up to $\frac{n-3}{4}$ corrupted workers.
Such number of corrupted workers~($m$) insures that the input for each run of $\mathcal{A}$ will have more than $2m$ workers as required, and there is also a majority of non-corrupted workers in the input to $TrimmedMean$.

We will follow the authors of this method and use  $\mathcal{A}$=Krum in the rest of the paper including our experiments.

\paragraph{No Defense}
In the experiments section we will use the name \textit{No Defense} for the basic method of averaging the parameters from all the workers, due to the lack of outliers rejection mechanism.

\section{Our Attack}
In previous papers \cite{krum,MeanMed,Bulyan}, the authors assume that the attacker will choose parameters that are far away from the mean, in order to hurt the accuracy of the model, for example by choosing parameters that are in the opposite direction of the gradient. Our attack shows that by consistently applying small changes to many parameters, a malicious opponent can perturb the model's convergence or backdoor the system. In addition, those defenses claimed to protect against an attacker which is omniscient, i.e. knows the data of all of the workers. We show that due to the normal distribution of the data, in case the attacker controls a representative portion of the workers, it is sufficient to have only the corrupted workers' data in order to estimate the distribution's mean and standard deviation, and manipulate the results accordingly. This observation enables our attack to work also for \textbf{non-omniscient} attacker, by estimating the properties of the entire population through the corrupted participants alone.

As mentioned above, the research in the field of distributed learning, including all defenses, assumes that the different parameters of all of the workers are i.i.d. and therefore expressed by normal distribution. We follow this assumption, hence in the rest of the paper the ``units" for attacking distributed learning models which we would like to maximize without being noticed are standard deviations ($\sigma$).

The outline of this section will go as follows: We will first analyze the range in which changes to the parameters will not be detected by \textit{TrimmedMean}, and upon choosing the maxima of this range the convergence is averted. Then we will point out a weakness in \textit{Krum} to be exploited by the attacker, claiming that the same parameters generated for the attack on TrimmedMean can also overcome Krum. This will lead to the conclusion that the same set of parameters will circumvent Bulyan as well because both of its building blocks were deceived. 
Finally, we show how the attacker can find the set of parameters within the same range that will introduce a backdoor to the system with only a minimal impact on accuracy for the original task.

The fact that the same set of parameters was used against all defenses is a strong advantage for this method: the attack will go unnoticed no matter which defense the server decides to choose, again rendering our attack more practical.

\subsection{Perturbation Range} \label{meanArndMed}
In the Trimmed Mean defense each parameter is calculated independently, so the problem can be considered as a one-dimension array, where each entry is the value given by a different worker. Clearly, if we will set the vicious value too far away from the mean, the malicious value will be discarded easily. We thus seek a range in which we can deviate from the mean without being detected. Since normal distribution is symmetric, the same value $z^{max}$ will set the lower and upper bounds for the applicable changes around the mean. 

What is the maximal change that can be applied by an attacker without being detected? In order to change the value produced by TrimmedMean, the attacker should control the median. This requires a majority of the workers, which can be attained by finding the minimal number $s$ of non-corrupted workers that needs to be ``seduced". The attacker will then use the properties of normal distribution, specifically the Cumulative Standard Normal Function $\phi(z)$, and look for value $z$ such that $s$ non-corrupted workers will probably reside further away from the mean. By setting all corrupted workers to values in the range ($\mu - z\sigma$, $\mu + z\sigma$), the attacker guarantees with high probability that those values will be the median and the many workers reporting the same value will cause it to withstand the averaging around the median in the second part of TrimmedMean. 

The exact steps for finding such a range are shown in Algorithm~\ref{alg:drift_attack} as part of the convergence prevention attack. 

\subsection{Overcoming Krum and Bulyan Defenses}
The output of Krum's process is only one chosen worker, and all of its parameters are being used while the other workers are discarded. It is assumed that there exists such a worker for which all of the parameters are close to the desired mean in each dimension. In practice however, where the parameters are in very high dimensional space, even the best worker will have at least a few parameters which will reside far from the mean.

To exploit this shortcoming, one can generate a set of parameters which will differ from the mean of each parameter by only a small amount. Those small changes will decrease the Euclidean Distance calculated by Krum, hence causing the malicious set to be selected. Experimentally, the attack on \textit{Trimmed Mean} was able to fool Krum as well.

An advantage when attacking Krum rather than Trimmed Mean is that only a few corrupted workers are required for the estimation of $\mu_j$ and $\sigma_j$, and only one worker needs to report the malicious parameters because Krum eventually picks the set of parameters originating from only a single worker.

Since Bulyan is a combination of \textit{Krum} and \textit{TrimmedMean}, and since our attack circumvents both, it is reasonable to expect that it will circumvent Bulyan as well. 

Nevertheless, Bulyan claim to defend against only up to 25\% of corrupted workers, and not 50\% like Krum and TrimmedMean. At first glance it seems that the $z^{max}$ derived for $m=25\%$ might not be sufficient, but it should be noted that the perturbation range calculated above is the possible input to \textit{TrimmedMean}, for which $m$ can reach up to 50\% of the workers in the $SelectionSet$ being aggregated in the second phase of Bulyan. 
Indeed, our approach is effective also against the Bulyan attack.

\subsection{Preventing Convergence}
With the objective of forestalling convergence, the attacker will use the maximal value $z$ that will circumvent the defense. The attack flow is detailed in Algorithm~\ref{alg:drift_attack}.

\begin{algorithm}[h]
\begin{algorithmic}[1]
\STATE {\bfseries Input:} \{$p_i : i \in CorruptedWorkers\}, n, m$
\STATE Set the number of required workers for a majority by:\\
\makebox[200pt]{$s = \lfloor \frac{n}{2} + 1 \rfloor - m $} \label{z_max_first}

\STATE Set (using \textit{z-table}):
$$z^{max} = \max_z \left(\phi(z) < \frac{n-s}{n}\right)$$

\FOR {$j \in [d]$}
	\STATE calculate mean ($\mu_j$) and  standard deviation ($\sigma_j$).\label{z_max_last}
    \STATE $(p_{mal})_j \gets \mu_j + z^{max} \cdot \sigma_j$
\ENDFOR

\FOR{$i \in$ CorruptedWorkers}
    \STATE $p_i \gets p_{mal}$ \; 
\ENDFOR

\caption{Preventing Convergence Attack}\label{alg:drift_attack}
\end{algorithmic}
\end{algorithm}

\textbf{Example:} If the number of malicious workers is 24 out of a total of 50 workers, the attacker needs to ``seduce" 2 workers ($\lfloor \frac{50}{2} + 1 \rfloor - 24 = 2$) in order to have a majority and set the median. $\frac{50-2}{50}= 0.96$, and by looking at the z-table for the maximal $z$ for which $\phi(z) < 0.96$ we get $z^{max}=1.75$\ifztable(Can be found in Table~\ref{tab:z_table})\fi. Finally, the attacker will set the value of all the malicious workers to $v = \mu + 1.75 \cdot \sigma$ for each of the parameters independently with the parameters' $\mu_j$ and $\sigma_j$. With high probability there will be enough workers with value higher than $v$, which will set $v$ as the median. 

In the experiments section we show that even a minor change of 1$\sigma$ can give the attacker control over the process at times.

\subsection{Backdooring Attack}
In section~\ref{meanArndMed}, we found a range for each parameter $j$ in which the attacker can perturb the parameter without being detected, and in order to obstruct the convergence, the attacker maximized the change inside this range. For backdooring attack on the other hand, the attacker seeks the set of parameters within this range which will produce the desired label for the backdoor, while minimizing the impact on the functionality for benign inputs.
To accomplish that, similar to \cite{vitaly2018backdoor}, the attacker will optimize for the model with the backdoor while minimizing the distance from the original parameters. This is achieved through the loss function, weighted by parameter $\alpha$ as follows:
\begin{equation}
Loss = \alpha \ell_{backdoor} + (1-\alpha)\ell_{\Delta}
\label{eq:Loss}
\end{equation}
where $\ell_{backdoor}$ is the same as the regular loss but trained on the backdoors with the attacker's targets instead of the real ones, and $\ell_{\Delta}$ to be detailed below is keeping the new parameters close to the original parameters.

For $\alpha$ too large, the parameters will significantly differ from the original parameters, thus being discarded by the defense mechanisms. Hence, the attacker should use the minimal $\alpha$ which successfully introduce the backdoor in the model.
Furthermore, the attacker can leverage the knowledge of $\sigma_j$ for each parameter, and instead of using any $L^p$ distance directly for $\ell_{\Delta}$, the difference between the parameters can be normalized in order to accelerate the learning:
\begin{equation}
\ell_{\Delta}= \sum_{j=1}^d {\left( \frac{NewParam_j - OldParam_j}{max(z^{max}\sigma_j, 1e-5)} \right)}^2  \label{eq:norm_loss}
\end{equation}

if $NewParam_j - OldParam_j$ is smaller than $z^{max}\sigma_j$, the new parameter is inside the valid range, so the ratio between them will be less than 1 and squaring it will reduce the value, which implies lower penalty. On the other hand, if $NewParam_j - OldParam_j$ is greater than $z^{max}\sigma$, the ratio is greater than 1 and the penalty increase quickly. 
Some $\sigma_j$ can happen to be very small, so values below $10^{-5}$ are being clamped in order to avoid division by very small numbers. This attack is detailed in Algorithm~\ref{alg:rep_attack}. 

\begin{algorithm}[h]
\begin{algorithmic}[1]

\STATE {\bfseries Input:} \{$p_i : i \in CorruptedWorkers\}, n, m$
\STATE Calculate $z^{max}, \mu_j$ and $\sigma_j$ as in Algo~\ref{alg:drift_attack}, lines~\ref{z_max_first}-\ref{z_max_last}. \;

\STATE Train the model with the backdoor, with initial parameters $\{\mu_j : j \in [d]\}$ and loss function described in equations~\ref{eq:Loss} and \ref{eq:norm_loss}.\; \label{malTrain}
\STATE $\mathcal{V} \gets$  final model parameters. \;

\FOR{$j \in [d]$}
	\STATE Clamp $v_j \in \mathcal{V}$ to the range $\mu_j \pm z_j^{max} \sigma_j$ using:
\ENDFOR
    \begin{equation*}
    (p_{mal})_j = max(\mu_j - z_j^{max} \sigma_j, min(v_j, \mu_j + z_j^{max} \sigma_j))
    \end{equation*}

\FOR{$i \in$ CorruptedWorkers}
    \STATE $p_i \gets p_{mal}$ \; 
\ENDFOR

\caption{Backdoor Attack}\label{alg:rep_attack}
\end{algorithmic}
\end{algorithm}

\section{Experiments and Results}
For our experiments we used PyTorch's \cite{pytorch} built in distribution package. 
In this section we describe the attacked models, and examine the impact on the models in the presence of different defenses for different $m$ and number of $\sigma$ ($z$).

\paragraph{Datasets} Following previous work \cite{MeanMed, TrimmedMean, Bulyan}, we consider two datasets. \textbf{MNIST} \cite{MNIST}, a hand-written digit identification dataset, and \textbf{CIFAR10} \cite{cifar10}, a classification task for $32\times32$ color images drawn from 10 different classes.

\paragraph{Models}
For both datasets, we follow the model 
architecture of the paper introducing the state of the art Bulyan defense \cite{Bulyan}.
For MNIST, we use a multi-layer perceptron with 1 hidden layer, 784 dimensional input (flattened $28\times28$ pixel images), a 100-dimensional hidden layer with ReLU activation, and a 10-dimensional softmax output, trained with cross-entropy objective. 
By using this structure, $d$ equals almost 80k. We trained the model for 150 epochs with \textit{batch size = 83}.
When neither attack nor defense are applied, the model reaches an accuracy of $96.1\%$ on the test set.

For CIFAR10 we use a 7-layer CNN 
 with the following layers: input of size 3072 ($32 \times 32 \times 3$); convolutional layer with kernel size: $3 \times 3$, 16 maps and 1 stride; max-pooling of size $3 \times 3$, a convolutional layer with kernel $4 \times 4$, 64 maps and 1 stride; max-pooling layer of size $4 \times 4$;  two fully connected layers  of size 384 and 192 respectively; and an output layer of size 10. We use ReLU activation on the hidden layer and softmax on the output, training the network for 400 epochs with a cross-entropy objective. In this setting $d \simeq 1M$. The maximal accuracy reached in this model with no corrupted workers is $59.6\%$, similar to the result obtained in \cite{Bulyan} for the same structure.

In both models we set the \textit{learning rate} and the \textit{momentum}  to be 0.1 and 0.9 respectively. We added L2 regularization with weight $10^4$ for both models. 
The training data was split between $n=51 = 4 \cdot m + 3$ workers, with $m = 12$ corrupted workers.

\ifrepurpose
\else
\subsection{Convergence Prevention}
\fi

 In Section~\ref{meanArndMed} we analyzed what is the maximal number of $\sigma$ away from $\mu$ that can be applied by our method, $z^{max}$. We showed in the example that when the total number of workers is 50,  the value of $z$ can be set to $1.75$, and all the corrupted workers will update each of their parameters values to $v =  \mu + 1.75\cdot\sigma$. Furthermore, when the total number of workers is greater than 50, $s$ still may equals 2 like before, but $\frac{n-s}{n}$ increases, causing an increase in the value of $z^{max}$ and further possible distance from the original mean. This can be intuitively explained given the fact that when $n$ increases, the chance for having outliers in the far tails of the normal distribution increases, and those tails are the ones to be seduced. In the following experiments, we tried to change the parameters by up to $1.5\sigma$, to leave room for inaccuracies with the estimation of $\mu_j$ and $\sigma_j$.
 
\paragraph{required $z$}
In order to learn how many standard deviations are required for impacting the network with the convergence attack, we trained the MNIST and CIFAR10 models in distributed learning settings four times, each time changing the parameters by $z= $ 0 (no change), 0.5, 1 and 1.5 standard deviations.  We did it for all the workers ($m=n$), on all parameters with no defense in the server. 
\begin{table}[hbtp]
\caption{\label{tab:acc_by_std} The maximal accuracy of MNIST and CIFAR10 models when changing all the parameters for all the workers. } 
\vskip 0.1in
\centering
\begin{tabular}{|l|c|c|c|c|}

\hline
\backslashbox{\textbf{Model}}{\textbf{$\sigma$}}                 & \textbf{0} & \textbf{0.5} & \textbf{1} & \textbf{1.5}\\ \hline
\textbf{MNIST}   & 96.1       & 89.0           & 82.4         & 77.8                   \\ \hline
\textbf{CIFAR10} & 59.6       & 28.4           & 20.9         & 17.5                 \\ \hline
\end{tabular}
\end{table}

As shown in Table~\ref{tab:acc_by_std}, it is sufficient to change the parameters by $1.5\sigma$ or even 1$\sigma$ away from the real average to substantially degrade the results. The table shows that degrading the accuracy of CIFAR10 is much simpler than MNIST, which is expected given the difference in nature of the tasks: MNIST is a much simpler task, so less samples are required and the different workers will quickly agree on the correct gradient direction, limiting the change that can be applied. While for the harder, more realistic classification task of CIFAR10, the disagreement between the workers will be higher, which can be leveraged by the malicious opponent.

\begin{figure}[t]
\centering
\includegraphics[width=0.48\textwidth]{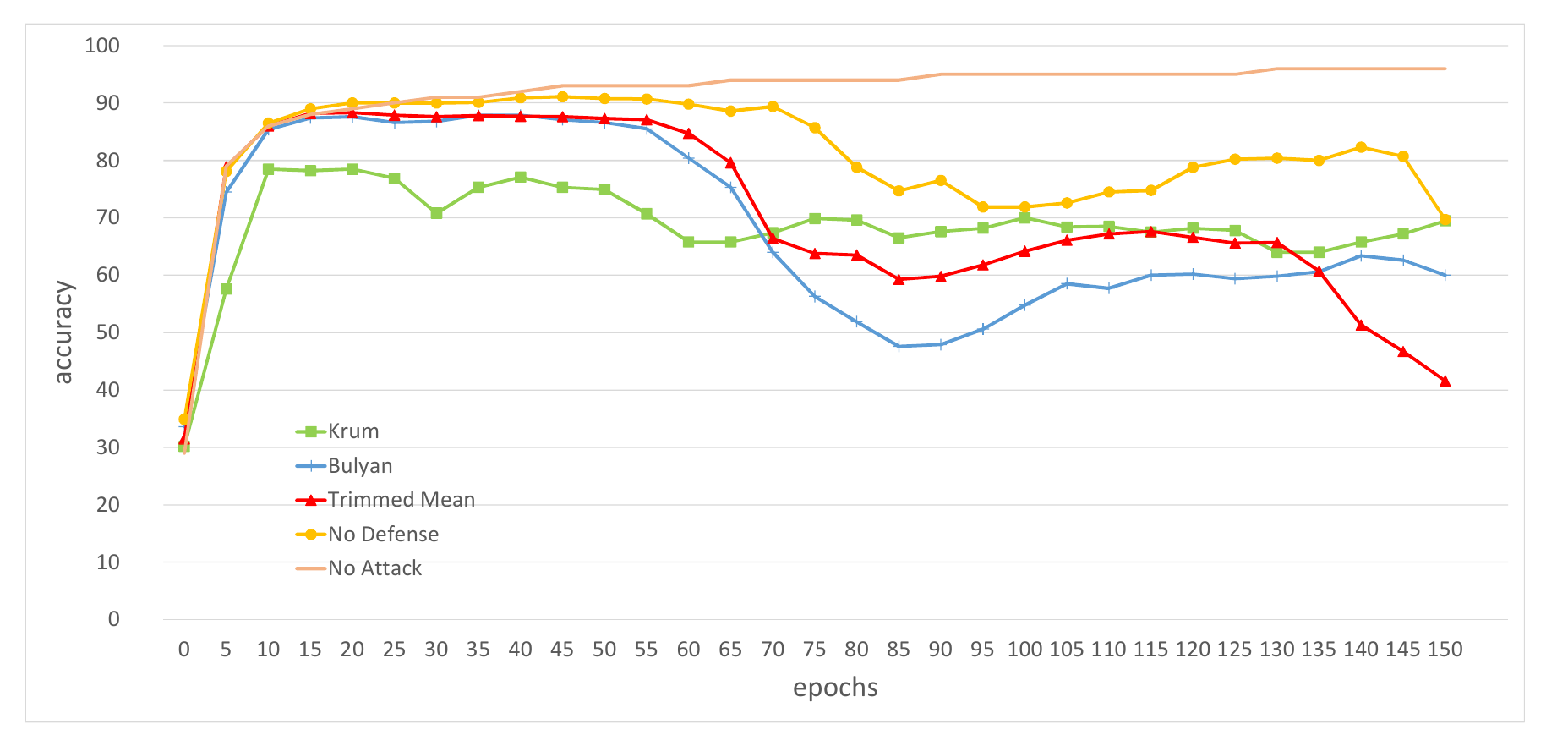}
\caption{Model accuracy on MNIST. $m=24\%$ and $z=1.5$. \textit{No Attack} is plotted for reference.}\label{attacks_2_std_graph}
\end{figure}

\paragraph{Comparing defenses}
We applied our attack against all defenses, and examined their resilience on both models. Figure \ref{attacks_2_std_graph} presents the accuracy of the MNIST classification model with the different defenses when the parameters were changed by $1.5\sigma$, over $m=12$ corrupted workers which is almost $24\%$. We also plotted the results when no attack is applied so the effect of the attack can clearly be seen. The attack is effective in all scenarios. The $Krum$ defense condition performed worst, since our malicious set of parameters was selected even with only 24\% of corrupted workers. \textit{Bulyan} was affected more than \textit{TrimmedMean}, because even though the malicious proportion was 24\%, it can reach up to 48\% of the \textit{SelectionSet}, which is the proportion used by TrimmedMean in the second stage of Bulyan. \textit{TrimmedMean} performed better than the previous two, because the malicious parameters were diluted by the averaging with many parameter sets coming from non corrupted workers.

Ironically but expected, the best defense strategy against this attack was the simplest aggregation rule of averaging without outliers rejection--- \textit{No Defense}. This is because the 1.5 standard deviations were averaged across all $n$ workers, 76\% of which are not corrupted, so the overall shift in each iteration was $1.5 * 0.24 = 0.36\sigma$, which only have a minor impact on the accuracy. It is clear however that the server cannot choose this aggregation rule because of the serious vulnerabilities it provokes. In case that circumventing \textit{No Defense} is desired, the attacker can compose a hybrid attack, in which one worker is dedicated to attack \textit{No Defense} with attacks detailed in earlier papers \cite{krum,MeanMed}, and the rest will be used for the attack proposed here.

\begin{figure}[tb]
\centering
\includegraphics[width=0.42\textwidth]{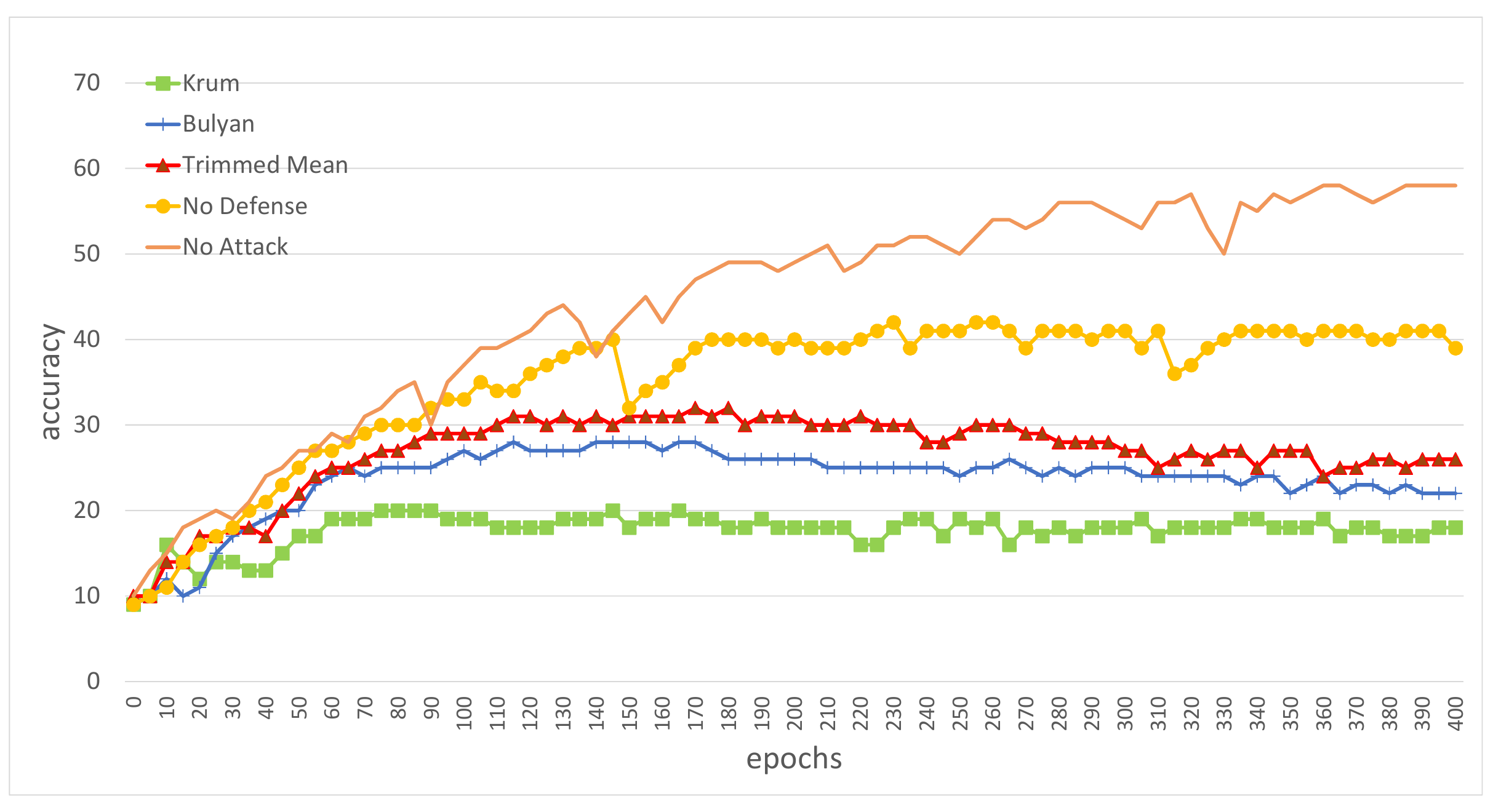}
\caption{Model accuracy on CIFAR10. $m=20\%$ and $z=1$. \textit{No Attack} is plotted for reference.}\label{attacks_1_std_graph_cifar}
\end{figure}

Experiment results on CIFAR10
are shown in Figure~\ref{attacks_1_std_graph_cifar}. Since fewer standard deviations can cause a significant impact on CIFAR10 (see Table~\ref{tab:acc_by_std}), we choose $m= 20\%$  corrupt workers, and change the parameters by only 1$\sigma$. Again, the best accuracy was achieved with the simplest aggregation rule, i.e. averaging the workers' parameters, but still the accuracy dropped by 28\%. \textit{Krum} performed worst again for the same reason with a drop of 66\%, \textit{Bulyan} dropped by 52\% and \textit{TrimmedMean} performed slightly better but still dropped by 45\%.

\begin{figure}[thb]
\centering
\includegraphics[width=0.48\textwidth]{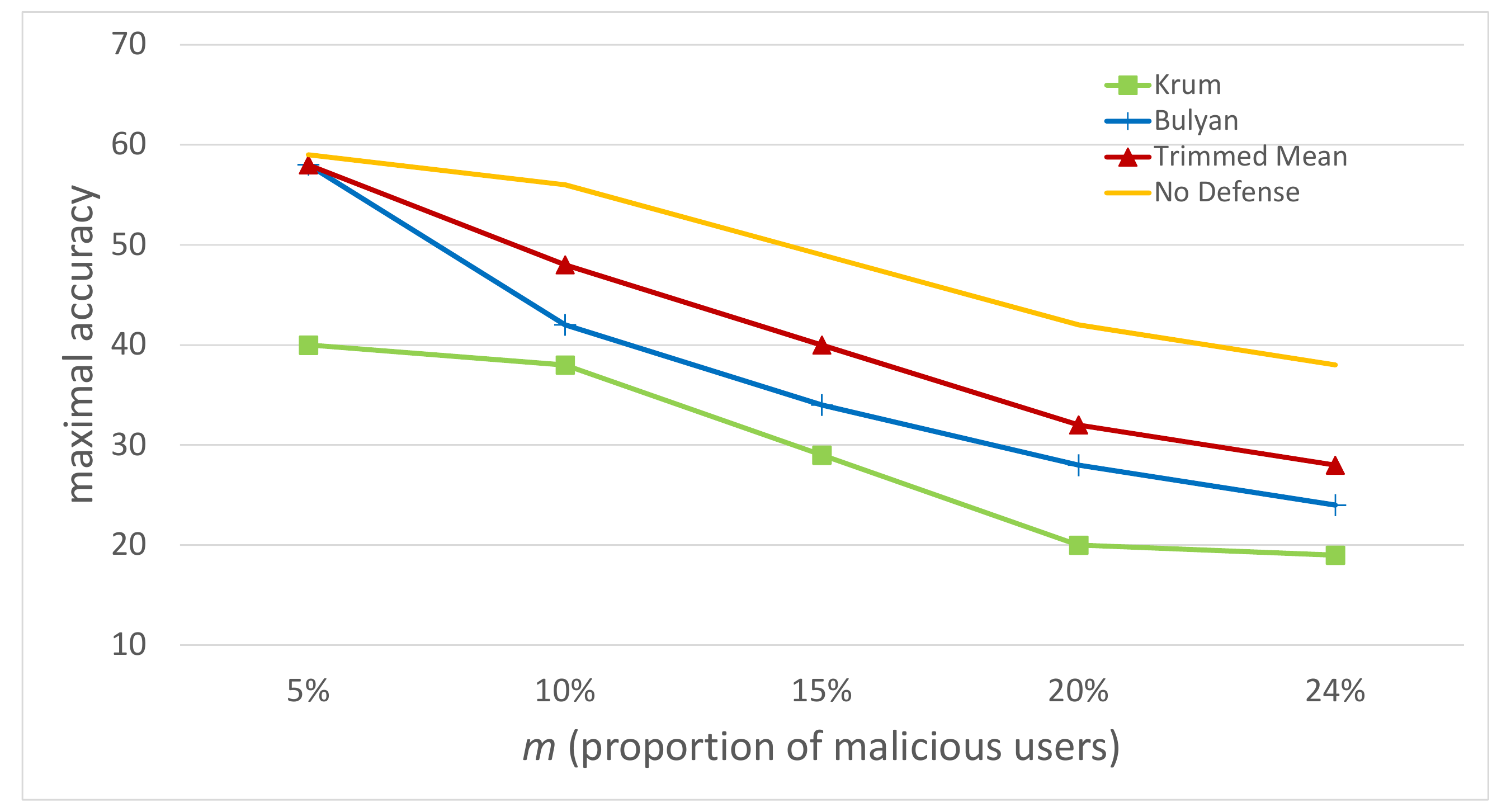}
\caption{Model accuracy with different proportion of corrupted workers ($m$) on CIFAR10. $z = 1$.}\label{attacks_mal_prop_graph}
\end{figure}

\paragraph{Proportion of malicious workers}
Figure~\ref{attacks_mal_prop_graph} shows the proportion of corrupted workers required to attack the training of CIFAR10 model. Since \textit{Bulyan} designed to protect against up to 25\% malicious workers, we tried to train the model with different $m$s up to that value, and tested how it affected the accuracy when the attacker changes all the parameters by $1\sigma$. One can see that Krum is sensitive even to a small amount of corrupted workers, thus even with $m = 5\%$ the accuracy drops by 33\%. The graph shows that as expected, as the proportion of corrupted workers grows, the model's accuracy decreases, but even 10\% can cause a major degradation with existing defenses other than not defending at all, which is not a realistic option.

\subsection{Backdooring}
As before, we set $n=51$ and $m=12$ (24\%). As a result of the attacker's desire not to interrupt the convergence for benign inputs, low $\alpha$ and $z$ (both 0.2) were chosen. After each round the attacker trained the network with the backdoor for 5 rounds. 
We set $\ell_{\Delta}$ 
according to Equation~\ref{eq:norm_loss} and set $\ell_{backdoor}$ to \textit{cross entropy} like the one used for the original classification.

\paragraph{Sample Backdooring} For the backdoor sample task, we chose each time one of the first 3 images from each training set (MNIST and CIFAR10) and take their desired backdoored targets to be $(y+1)\mod|Y|$ where $y$ is the original label and $|Y|$ is the number of classes.

Results are presented in Table~\ref{table:backdooring}. Throughout the process, the network produced the malicious target for the backdoor sample in more than 95\% of the time, including specifically the rounds where the maximal overall accuracy was achieved. As can be seen, for a simple task such as MNIST where the network has enough capacity, the network succeeded to incorporate
the backdoor with less than 1\% drop in the overall accuracy. The results are similar across the different defenses by cause of the low $z$ being used. For CIFAR10 however, where the convergence is difficult even without the backdoor for the given simple architecture, the impact is more visible and reaches up to 9\% degradation.

\begin{table}[htb]
\caption{\label{table:backdooring} \textbf{Backdoor Sample Results.} The maximal accuracy of MNIST and CIFAR10 models with a backdoor sample. $n=51, m=24\%, z=\alpha=0.2$. The results with no backdoor introduction are also presented for comparison.
}
\vskip 0.1in

\centering
\begin{tabular}{|l|c|c|}
\hline
\backslashbox{\textbf{Defense}}{\textbf{Model}}     & \textbf{MNIST} & \textbf{CIFAR10} \\ \hline
\textbf{No Attack}                                  & 96.1           & 59.6             \\ \hline
\textbf{No Defense}                                 & 95.4           & 58.4             \\ \hline
\textbf{Trimmed Mean} & 95.4           & 57.9             \\ \hline
\textbf{Krum}         & 95.3           & 54.4             \\ \hline
\textbf{Bulyan}       & 95.3           & 54.2             \\ \hline
\end{tabular}
\end{table}

\paragraph{Pattern Backdooring}
For the backdoor pattern attack, the attacker randomly samples 1000 images from the datasets on each round, and set their upper-left 5x5 pixels to the maximal intensity (See Figure~\ref{objectives} for examples). All those samples were trained with $target = 0$. For testing the same pattern was applied to a different subset of images.

 Table~\ref{table:backdooring_pattern} lists the results. Similar to the results for backdoor sample case, MNIST perfectly learned the backdoor pattern with a minimal impact on the accuracy for benign inputs on all defenses except for \textit{No Defense} where the attack was again diluted by the averaging with many non-corrupted workers, and yet the malicious label was selected for non-negligible 36.9\% of the samples. For CIFAR10 the accuracy is worse than with the backdoor \textit{sample}, with a 7\% (\textit{TrimmedMean}), 12\% (\textit{Krum}) and 15\% (\textit{Bulyan}) degradation, but the accuracy drop for benign inputs is still reasonable and probably unsuspicious for an innocent server training for a new task without knowing the expected accuracy. For each of the three defenses, more than 80\% of the samples with the backdoor pattern were classified maliciously. 

It is interesting to see that \textit{No Defense} was completely resilient to this attack, with only a minimal degradation of 1\% and without mis-classifying samples with the backdoor pattern. However, on a different experiment on MNIST with higher $z$ and $\alpha$ (1 and 0.5 respectively), the opposite occur, where No Defense reached 95.6 for benign inputs and 100\% on the backdoor, while other defenses did not perform as well on the benign inputs. Another option for circumventing \textit{No Defense} is 
dedicating one corrupted worker for the case that \textit{No Defense} is being used by the server, and use the rest of the corrupted workers for the defense-evading attack.

\begin{table}[hbt]
\caption{\label{table:backdooring_pattern} \textbf{Backdoor Pattern Results.} The maximal accuracy of MNIST and CIFAR10 models with backdoor pattern attack. $n=51, m=24\%, z=\alpha=0.2$. The results with no backdoor introduction are also presented for comparison. Results are presented for legitimate inputs (benign) and images with the backdoor pattern. 
}
\vskip 0.1in

\begin{tabular}{|l|c|c|c|c|} 
\hline
\multirow{2}{*}{} & \multicolumn{2}{c|}{\textbf{MNIST}} & \multicolumn{2}{c|}{\textbf{CIFAR10}}  \\ 
\cline{2-5}
                                                                 & \thead{Benign} & \thead{Backdoor} & \thead{Benign} & \thead{Backdoor~}   \\ 
\hline
\textbf{No Attack}                                               & 96.1            & -~                & 59.6            & -~                   \\ 
\hline
\textbf{No Defense}                                              & 96.0              & 36.9              & 59.1            & 7.3                  \\ 
\hline
\textbf{\begin{tabular}[c]{@{}l@{}}Trimmed \\ Mean\end{tabular}} & 95.3            & 100.               & 55.6            & 80.7                 \\ 
\hline
\textbf{Krum~}                                                   & 95.2            & 100.               & 52.5            & 95.1                 \\ 
\hline
\textbf{Bulyan}                                                  & 95.3            & 99.9              & 51.9            & 84.3                 \\
\hline
\end{tabular}
\end{table}

\section{Conclusions}
We present a new attack paradigm, in which by applying limited changes to many parameters, a malicious opponent may \ifrepurpose\else \textbf{interfere with or backdoor} \fi the process of \textit{Distributed Learning}. Unlike previous attacks, the attacker does not need to know the exact data of the non-corrupted workers (being \textbf{non-omniscient}), and it works even on i.i.d. settings, where the data is known to come from a specific distribution. The attack evades all existing defenses. Based on our experiments, a variant of \textbf{TrimmedMean} is to be chosen among existing defenses, producing the best results for convergence attack excluding the choice of na\"ive averaging, which is obviously vulnerable to  other simpler attacks.
\bibliographystyle{apalike}
\bibliography{refs}

\vspace{12pt}

\end{document}